\documentclass[runningheads]{llncs}
\usepackage[T1]{fontenc}
\usepackage{graphicx}
\usepackage{booktabs}
\usepackage[misc]{ifsym}
\usepackage{subcaption}

\usepackage{mwe}
\usepackage[utf8]{inputenc} 
\usepackage[T1]{fontenc}    
\usepackage{hyperref}       
\usepackage{url}            
\usepackage{amsfonts}       
\usepackage{nicefrac}       
\usepackage{microtype}      
\usepackage{xcolor}         
\usepackage{amsmath,amssymb,amsfonts}
\usepackage{bm}
\usepackage{listings}
\usepackage{makecell}
\usepackage{comment}

\def\code#1{\texttt{#1}}

\usepackage{siunitx}       
\usepackage{tabularx}      
\usepackage{threeparttable}
\usepackage{multirow}      
\usepackage{xspace}
\usepackage{orcidlink}

\begin{document}

\title{MotifGen: Spatiotemporal interpolation of misaligned satellite images via multi-source generative modeling, in an application to tropical cyclones}

\titlerunning{Generative multi-source spatiotemporal interpolation of tropical cyclones}

\author{Clément Dauvilliers\inst{1}\orcidlink{0009-0006-2292-0926} \and
Claire Monteleoni\inst{1,2}\orcidlink{0000-0002-9488-0517}}

\authorrunning{C. Dauvilliers and C. Monteleoni}

\institute{INRIA, Paris, Île-de-France, France \and
University of Colorado Boulder, Boulder, CO, USA.}

\maketitle              

\begin{abstract}
Microwave satellite imagery plays a crucial role in monitoring tropical cyclone precipitation and intensity worldwide, but suffers from long revisit times, potentially missing rapid storm evolution phases. While this raises the need for an interpolation method, it is made challenging by the high level of heterogeneity of microwave data coming from different instruments. In this work, we introduce the first generative model that can be applied to multiple geospatial sources that change across samples, occur at irregular time intervals, are misaligned geographically, and come from instruments with varying characteristics. We apply this model to the case of spatio-temporal interpolation of tropical cyclone microwave images from other microwave and infrared instruments. We train using a self-supervised task in which a random source is masked and reconstructed, and show that it leads to a significant decrease in Continuous Ranked Probability Score over supervised training. We show a further improvement by combining infrared and microwave data compared to microwave only. Using these improvements, the generative model produces an ensemble mean on par with that of a deterministic model, while generating a power spectrum significantly closer to that of true observations. To the best of our knowledge, this is the first generative model that interpolates microwave images of cyclones by combining multiple microwave instruments and infrared observations at irregular time intervals.

\keywords{Multi-source machine learning \and Satellite imagery \and Generative models \and Tropical cyclones.}
\end{abstract}

\section{Introduction}
Tropical cyclones (TCs) are among the most destructive extreme weather events globally, both in terms of human fatalities and economic loss. Satellite imagery is a core ingredient in their monitoring, including in estimating and forecasting their intensities, sizes, and trajectories \cite{veldenConsensusApproachEstimating2020,duongObjectiveSatelliteMethods2023,dauvilliersMoTiFSelfsupervisedModel2025,youPredictingTropicalCyclone2025,zhuoPhysicsAugmentedDeepLearning2021}. Among the products used to observe tropical cyclones, only geostationary satellites can provide near-continuous observations of cyclones, but these are limited to visible and infrared imaging \cite{duongObjectiveSatelliteMethods2023}. Other types of measurement, such as microwave images, are only available when a satellite equipped with an adapted sensor orbits over a storm, leading to high revisit times that can often miss crucial evolution phases \cite{haynesAidingTropicalCyclone2024,mengSimulatingTropicalCyclone2022}. This raises the need for methods that can interpolate observations of tropical cyclones for non-stationary satellites, in particular microwave images \cite{duongObjectiveSatelliteMethods2023}.

\begin{figure}[!ht]
    \centering
    \begin{subfigure}[b]{0.9\textwidth}
        \centering
        \includegraphics[width=\textwidth]{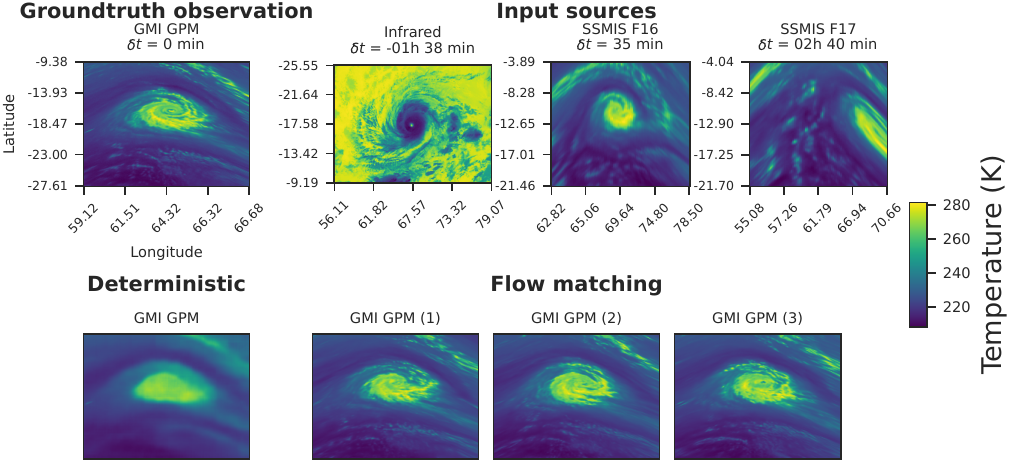}
        \caption{Inputs, target and reconstructions of the 37GHz microwave channel. Top left: target observation. Top right group: input sources, with $\delta t$ indicating the time difference between the target and each input source. Bottom left: reconstruction from a deterministic model trained to optimize the RMSE. Bottom right: reconstructions generated by a flow matching model with different input noises.}
        \label{fig:det_vs_fm}
    \end{subfigure}

    \vspace{1em} 
    
    \begin{subfigure}[b]{0.45\textwidth}
        \centering
        \includegraphics[width=\textwidth]{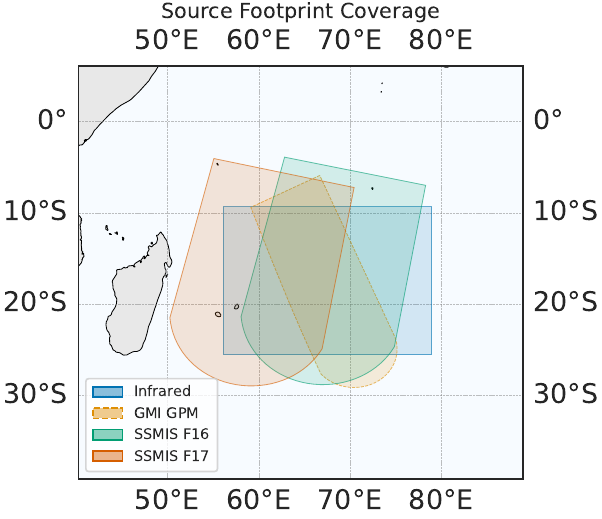}
        \caption{Footprint of the satellite images (inputs and target) in the sample displayed in Figure \ref{fig:det_vs_fm}.}
        \label{fig:footprint}
    \end{subfigure}
    
    \caption{Example of interpolation for a deterministic model versus a generative one.}
    \label{fig:reconstruction_example}
\end{figure}

However, satellite imagery is a particular modality in machine learning, as the large diversity of instruments leads to highly heterogeneous data \cite{rolfPositionMissionCritical2024}. Interpolating microwave images of tropical cyclones is no exception, mainly as microwave images are provided by a constellation of satellites rather than a lone instrument. The consequence is that, for a given cyclone, microwave observations occur at unpredictable times, depending on when one of the constellation's satellites orbits over the storm's location. This implies that the time interval between two successive observations is irregular, and that within any given time window the number of available microwave images isn't known in advance. In addition, the observations are generally geographically misaligned, as tropical cyclones most often travel significant distances between two successive microwave images. Finally, the microwave sensors within the constellation present important differences in their characteristics, such as the measured frequency, viewing angle, spatial geometry, resolution, and swath width, among others. This means that models trained on data from a single sensor do not generally transfer well to other instruments \cite{pfreundschuhGPROFNNNeuralnetworkbasedImplementation2022,sambathUnsupervisedDomainAdaptation2024}. All of these aspects (irregular time intervals, geographical misalignment, and varying characteristics) make microwave images of tropical cyclones a particularly heterogeneous type of data, thus requiring adapted machine learning methods and architectures.

In addition to the challenge of heterogeneity, real observations do not generally contain the full information required to interpolate images of a cyclone at another time and place, making the task inherently probabilistic. Consequently, deterministic models for example trained to optimize the Root Mean Square Error (RMSE) output blurry, nonphysical predictions \cite{dauvilliersMoTiFSelfsupervisedModel2025} - an example is shown in Figure \ref{fig:det_vs_fm}. This raises the need to switch to generative methods such as flow matching \cite{lipmanFlowMatchingGenerative2023}, which allow to sample from an approximation of the target distribution.\\

In this work, we address these two challenges by proposing a generative framework to perform spatio-temporal interpolation of heterogeneous satellite observations. We use this framework to interpolate high-resolution satellite images of tropical and extra-tropical cyclones: our model receives as input observations from a large pool of microwave and infrared sensors, and reconstructs images from the GPM Microwave Imager (GMI), the reference instrument for the GPM constellation. As the amount of GMI images to use as training target is limited, we train our model using a self-supervised reconstruction task, and show that it leads to a significant improvement in Continuous Ranked Probability Score (CRPS) over training in a supervised manner. We show a further gain by including infrared images on top of microwave images in the model's input. We validate our generative model by comparing its ensemble mean against a deterministic version adapted from \cite{dauvilliersMoTiFSelfsupervisedModel2025}. Our model obtains comparable results on deterministic metrics (RMSE, MAE, MAPE), with a power spectrum significantly closer that of the groundtruth.
The summary of our contributions is the following:
\begin{itemize}
    \item We introduce a generative model that can be applied to multiple geospatial sources that change across samples, occur at irregular time intervals, are misaligned geographically, and come from instruments with varying characteristics.
    \item We apply this model to perform spatio-temporal interpolation of high-resolution microwave images of tropical and extra-tropical cyclones. To the best of our knowledge, this is the first generative model that produces microwave images of cyclones by combining multiple microwave instruments and infrared observations.
    \item We show that by leveraging self-supervised training and including infrared data in the input, our model reaches a better Root Mean Squarer Error than a deterministic model trained in the same conditions, while producing a significantly more realistic power spectrum.
\end{itemize}
This paper is organized as follows: Section \ref{sec:related_work} introduces the related work; Section \ref{sec:methods} details the multi-source generative framework; Section \ref{sec:experiments} describes the experiments and associated results; finally \ref{sec:conclusion} concludes the paper.

\section{Related Work}
\label{sec:related_work}
In this section, we discuss recent works that specifically study geospatial data through the use of generative modeling and multi-source machine learning.

\paragraph{Generative models for meteorological data} The success of generative models in recent years in natural images has naturally led to their widespread adoption for geospatial data in turn. In the context of meteorological data, generative models have been used with great success for global weather forecasting \cite{couaironArchesWeatherArchesWeatherGenDeterministic2024,priceProbabilisticWeatherForecasting2025}, atmospheric downscaling \cite{tuMODSMultisourceObservations2025,mardaniResidualCorrectiveDiffusion2025,merizziWindSpeedSuperresolution2024}, and precipitation nowcasting \cite{aspertiPrecipitationNowcastingGenerative2023,liPrecipitationNowcastingUsing2024}, among others. In the specific case of microwave data, \cite{sambathUnsupervisedDomainAdaptation2024} used a CycleGAN to perform domain adaptation of microwave images from the SSMIS F18 sensor satellite to GMI images; \cite{guilloteauGenerativeDiffusionModel2024} used a diffusion model to estimate precipitation maps from SSMIS F17 and infrared observations. In the specific case of tropical cyclones, \cite{hanGeoGMIGenerativeAdversarial2026} use a GAN to generate GMI images from infrared data. While these papers highlight the usefulness of generative models to generate microwave and precipitation data, they only generate images over the same area as their inputs. In contrast, we perform spatio-temporal interpolation, meaning we train a model to reconstruct images over a different area and at different times from the input sources, which is required to track a cyclone whose location changes with time. Besides, \cite{sambathUnsupervisedDomainAdaptation2024,guilloteauGenerativeDiffusionModel2024,hanGeoGMIGenerativeAdversarial2026} use a single, fixed source as input, while we suggest to use a flexible number of input sources from multiple different instruments.

\paragraph{Multi-source self-supervised models for geospatial data} Multi-modal machine learning for geospatial and especially remote sensing data is an active field. A large part of it is foundation models that are trained in a self-supervised manner to take advantage of the large amount of publicly available unlabeled data \cite{congSatMAEPretrainingTransformers,klemmerSatCLIPGlobalGeneralPurpose2024,astrucOmniSatSelfsupervisedModality2025}. Self-supervised learning pairs well with multi-modal geospatial problems: by gathering multiple sources over a common geographical area, one can exploit their correlations to train a latent representation \cite{jakubikTerraMindLargeScaleGenerative2025}. In the case of meteorological data, MODS \cite{tuMODSMultisourceObservations2025} recently introduced a model to downsample meteorological states, using multiple sources as input and a self-supervised reconstruction task. While MODS relies on a fixed set of sources, MoTiF \cite{dauvilliersMoTiFSelfsupervisedModel2025} propose a self-supervised architecture and self-supervised training task adapted to multi-source geospatial data, and use those to interpolate microwave images of tropical cyclones. This work is essentially an extension of \cite{dauvilliersMoTiFSelfsupervisedModel2025}, with the following differences: (1) we modify the framework to train a flow matching model instead of optimizing the RMSE; (2) we use a combination of microwave and infrared as input data.

\section{Methods}
\label{sec:methods}
This section presents the details of our methodology. Subsection \ref{sec:problem} first formulates the problem; then \ref{sec:flow_matching} gives an overall view of flow matching and how we employ it the context of this work. Finally, \ref{sec:architecture} details the architecture we use to process multi-source data.

\subsection{Problem Formulation}
\label{sec:problem}

\subsubsection{Definition of an observation from a source}\label{sec:source_item} Let $S_i$ be a source (e.g. one the microwave instruments). An element $\bm{x}$ from $S_i$ includes the following components: the pixels, as an image $\bm{v}_i\in\mathbb{R}^{C\times H\times W}$; the latitude and longitude at each pixel, as an array of shape $\bm{c}\in\mathbb{R}^{2\times H\times W}$; the time $t$ associated with the element; and a vector of characteristic variables $\bm{s}\in\mathbb{R}^5$. These characteristics are the instantaneous field-of-view (IFOV) at nadir along and across track, at the edges along and across track, and the observing frequency in GHz. Finally, $\bm{x}$ also includes a land-sea mask and a binary availability mask valued at zero at missing pixels. In practice, the dimensions $H$ and $W$ of the pixels, coordinates and masks are specific to each observation.

\subsubsection{Definition of a sample}\label{sec:sample} Let $\mathcal{S}=\{S_1,S_2,...,S_{|\mathcal{S}|}\}$ be the set of all sources in the dataset. A sample $\bm{X}$ is defined as a set $\bm{X}=(\bm{x}_0,\bm{x}_1,\bm{x}_2,...,\bm{x}_K)$ of observations from a subset of $\mathcal{S}$. A source $S_i$ may be present or not in $\bm{X}$, and can appear multiple times. We only consider a sample for training or evaluation if it includes at least two observations, so that one can be used as target (either for training or evaluation) and the other as conditioning for the generation. Every sample is defined around its reference observation $\bm{x}_0$, which itself defines the sample's reference time $t_0$. Using the observation times $t_0,t_1,t_2,...,t_K$, we compute the relative time differences
\begin{equation*}
    \Delta t_{i} = \frac{t_i - t_0}{\Delta t_{max}}
\end{equation*}
where $\Delta t_{max}$ is the maximum time difference between two observations that are within the same sample, and which is a hyperparameter of the experiment. The intuition behind this choice is that the time difference between the input sources and between the input and output sources is valuable for the interpolation task, while absolute times would be over-specific and could lead to overfitting. Dividing by the model's assimilation window normalizes the time differences to stay within $[0,1]$. 

\subsubsection{Reconstruction task}\label{sec:task}
Let $\bm{X}=(\bm{x}_0,\bm{x}_1,\bm{x}_2,...,\bm{x}_K)$ be a sample defined as above, with $\bm{x}_0$ being the reference observation. Our objective is to learn a model that reconstructs the reference image's pixels, conditioned on the other sources and the reference source's coordinates, time and characteristics:
\begin{equation*}
    F_\theta(\bm{x}_1,\bm{x}_2,...,\bm{x}_3,\bm{c}_0,\bm{s}_0, \Delta t_0) \approx \bm{v}_0
\end{equation*}
In other words, the model generates a synthetic image with the characteristics of the reference source, at a requested time and over a requested geographic area, conditioned on a set of input sources to guide its generation.

\subsection{Flow matching}
\label{sec:flow_matching}
Flow matching \cite{lipmanFlowMatchingGenerative2023} is a form of generative modeling in which a starting distribution $p_{init}$ that can be easily sampled (often a standard Gaussian) is transported to a data distribution $p_{data}$ via a flow function $\psi_r: \mathbb{R}^d\times[0,1] \rightarrow\mathbb{R}^d$. 

\begin{figure}
    \centering
    \includegraphics[width=0.5\linewidth]{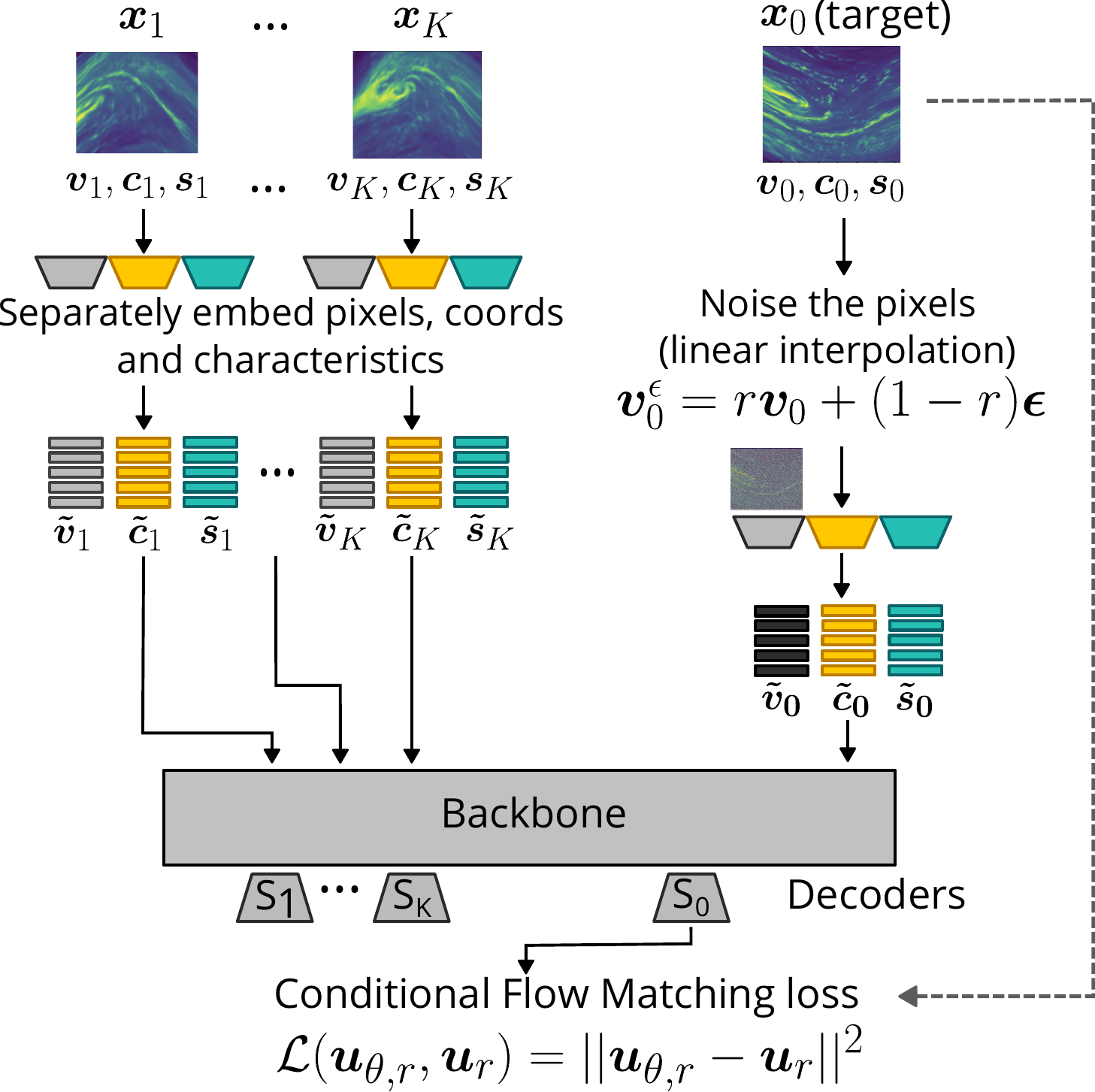}
    \caption{Overall training pipeline.}
    \label{fig:pipeline}
\end{figure}

To do so, flow matching defines the flow via an ordinary differential equation (ODE):
\begin{equation}
\label{eq:fm_ode}
    \frac{d}{dr}\psi(x_0) = u_r(\psi(x_0)) \quad \text{for }\ r\in[0,1], \quad \text{s.t.}\quad \psi_0(x_0) = x_0
\end{equation}
where $u_r: \mathbb{R}\times[0,1]\rightarrow \mathbb{R}^d$ is a marginal \textit{velocity field} such that if $X_0 \sim p_{0}=p_{init}$ and $X_r$ is a solution of Equation \ref{eq:fm_ode}, then $X_1 \sim p_1=p_{data}$. The model is trained to approximate the velocity field, i.e. $u^\theta_r \approx u_r$. Since $u_r$ is intractable, flow matching relies on the \textit{marginalization trick} (Theorem 1 in \cite{lipmanFlowMatchingGenerative2023}), which states that one can equivalently learn conditional velocity fields $u_r(x |x_0,x_1)$. This is done by optimizing the conditional flow matching loss:
\begin{equation*}
    \mathcal{L}_{CFM}(\theta) = \mathbb{E}_{r\sim p_r,X_0\sim p_{init},X_1\sim p_{data}} ||u^\theta_r(X_r)-u_r(X_r|X_0,X_1)||^2
\end{equation*}
In this work, we use the CondOT path \cite{lipmanFlowMatchingGenerative2023}, which simply defines the conditional flow as a linear interpolation between a noise point and a data point:
\begin{equation*}
    \psi_r(x_0|x_1)=rx_1+(1-r)x_0
\end{equation*}
which corresponds to the conditional velocity field $u_r(x_r|x_0,x_1)=x_1-x_0$. In our case, we replace the target source's pixels with a version corrupted by standard Gaussian noise, while leaving the input sources unchanged. The training process is described in Figure \ref{fig:pipeline}. During training, we sample $r\sim p_r$ following a lognormal distribution \code{lognormal(0.0, 1.0)}, following the recommendations in \cite{esserScalingRectifiedFlow2024}. Inferring with a trained model is done by sampling $x_0\sim \mathcal{N}_{d_{pixels}}(0,I)$ and numerically solving Equation \ref{eq:fm_ode} via the 1-step Euler method:
\begin{equation*}
    x^\theta_{r+h} = x^\theta_r+hu^\theta_r(x_r^\theta)
\end{equation*}
where $h=\frac{1}{N}$ is the solving step size. We set $N=25$ in our experiments.

\subsection{Multi-source architecture}
\label{sec:architecture}
We use an architecture based on MoTiF \cite{dauvilliersMoTiFSelfsupervisedModel2025}, which we adapt for flow matching. It is similar to the widely used Diffusion Transformer \cite{peeblesScalableDiffusionModels2023}, with significant differences designed to handle multiple misaligned geospatial sources. The architecture uses three separate embedding spaces for the pixels, spatio-temporal coordinates and characteristics of the source. The coordinates embedding serves as positional encoding: since the sources correspond to different geographical areas, a traditional encoding based on the position of a patch within the image would loose any sense when computing the attention across sources. For example, the center of the cyclone could be in the top-left corner of one observation but in the bottom-right of another, depending on the satellites' orbits. Instead, we use the spatio-temporal coordinates (latitudes, longitudes, time) as positional encoding, as was done in \cite{dauvilliersMoTiFSelfsupervisedModel2025}, since those have a common meaning across sources.

\subsubsection{Embedding layers} 
The model possesses two embedding layers, for microwave images and for infrared data, which have separate weights the same functioning. To begin with, the channels are concatenated with the land-sea mask and the availability mask. The result $\bm{v}\in\mathbb{R}^{(C+2)\times H\times W}$ is then cut into square patches and embedded to the pixel embedding dimension $d_{pixels}$ following the common ViT patchifying process \cite{dosovitskiyImageWorth16x162021}.

Regarding the coordinates, the latitudes, longitudes and time are concatenated, and then converted into sine-cosine features before being embedded as patches to a dimension $d_{coords}$. This follows the common usage in (vision) transformers for positional embedding \cite{vaswaniAttentionAllYou2023,dosovitskiyImageWorth16x162021,esserScalingRectifiedFlow2024}. Besides, this transformation accounts for the periodicity of latitudes and longitudes.

Finally, the embedding layers also produce a conditioning sequence, which is also spatial and has the same shape as the embedded pixels and coordinates. It is built by first embedding the land-sea mask via patches to a dimension $d_{cond}$. Then, the source characteristics vector and an availability flag are projected to $d_{cond}$ via a linear layer.  Finally, the flow matching time step $r\in[0,1]$ is embedded via sine-cosine features, as is commonly done for diffusion / flow matching models \cite{esserScalingRectifiedFlow2024}. The embedded land-sea mask, availability flag, characteristics and flow matching step are then summed together to obtain the final conditioning sequence.

The availability flag is a scalar whose value is 1.0 for input sources, 0.0 for the target source whose pixels are masked, and -1.0 for missing sources. As a source may be present in one sample but absent from another one within the same mini-batch, empty tensors are inserted to maintain the same structure across samples. The availability flag is added in the conditioning to carry the information over whether each source is available or missing.

\subsubsection{Backbone} The backbone is similar to that of DiT \cite{peeblesScalableDiffusionModels2023}, in the sense that it is a chain of blocks that each contain attention layers, followed by a Multi-Layer Perceptron (MLP), each wrapped in an adaptive conditional layer normalization (adaLN). However, instead of using a single self-attention layer across all sources, each block contains a source-wise spatial self-attention layer followed by a cross-source attention layer. 

\paragraph{Source-wise self-attention} The self-attention layers compute the attention within each source individually using 2D Swin windows \cite{liuSwinTransformerHierarchical2021}. However, as mentioned in \ref{sec:architecture}, we use the embedded spatio-temporal coordinates as relative positional encoding \cite{shawSelfAttentionRelativePosition2018}. For each source $S_i$, the embedded patches of pixels are projected to a triplet $(\bm{Q}_{pixels},\bm{K}_{pixels},\bm{V}_{pixels})$ of queries, keys and values of dimension $d_{pixels}$. The spatio-temporal coordinates are in turn projected to another pair of queries and keys $(\bm{Q}_{coords},\bm{K}_{coords})$, of dimension $d_{coords}$. The attention-weighted values are then computed as
\begin{equation}
    \label{eq:attention}
    \bm{V}_{\text{out}}= \text{Softmax}\left( \frac{\bm{Q}_{pixels}\bm{K}^T_{pixels}}{\sqrt{d_{pixels}}} + \alpha \frac{\bm{Q}_{coords}\bm{K}^T_{coords}}{\sqrt{d_{coords}}} \right)\bm{V_{pixels}}
\end{equation}
where $\alpha$ is a learnable parameter that lets the model adapt the importance of the coordinates term.

\begin{figure}
    \centering
    \includegraphics[width=1.0\linewidth]{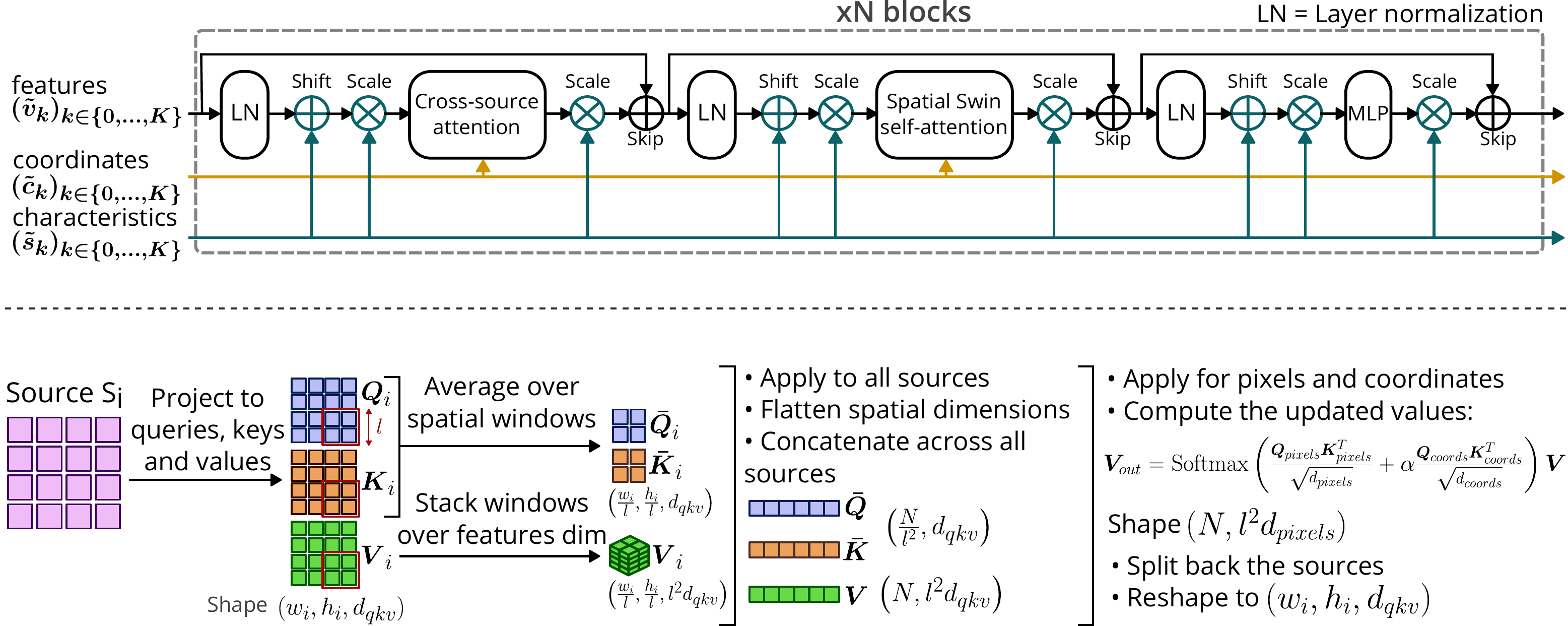}
    \caption{Diagram of the architecture. Top: overall view of the backbone. Bottom: scheme of the cross-source attention.}
    \label{fig:architecture}
\end{figure}

\paragraph{Cross-source attention } As the sources are geographically misaligned, 3D Swin attention \cite{biAccurateMediumrangeGlobal2023} or axial attention methods \cite{couaironArchesWeatherArchesWeatherGenDeterministic2024} cannot be applied to let information travel across sources. On the other hand, directly computing an attention matrix over the full sequence of all patches from all sources would be prohibitively costly. While \cite{dauvilliersMoTiFSelfsupervisedModel2025} proposed a cross-source attention attention based on anchor points for a deterministic model, we found that it impaired performance for our generative task, likely due to compressing the embeddings within the attention layer. For this reason, we design a new cross-source attention layer, which lets information travel between any areas of any source, while reducing the cost by a constant factor, and without compressing the values. Its process is described in Figure \ref{fig:architecture}.

First, the projected queries and keys are split into 2D spatial square windows of side $l$, and averaged to obtain one query and key per window. The values are also split into the same windows, then stacked along the feature dimension. This process is repeated for each source. The queries, keys and values thus obtained are concatenated into sequences of length $\frac{N}{l^2}$ for queries and keys, and $N$ for the values, where $N$ is the total number of patches across all sources. The same is done to convert the embedded coordinates into keys and queries, resulting in two pairs $(\bm{Q}_{pixels},\bm{K}_{pixels}), (\bm{Q}_{coords},\bm{K}_{coords})$. The attention-weighted values are then computed using Equation \ref{eq:attention}, resulting in a set of updated values of shape  $(N, l^2d_{pixels})$. These then go through the reverse process of being unstacked and reshaped into the original shape of each source. Through the averaging of the keys and queries, this layer has a memory cost of $\mathcal{O}(N^2/l^2)$, which while it remains quadratic, still reduces by the cost by 16 times with windows of length $l=4$, compared to full attention. In practice, the computation of the attention matrix is done using multiple attention heads, as is commonly done in transformers \cite{vaswaniAttentionAllYou2023}.

\section{Experiments}
\label{sec:experiments}
This section describes the details of the experiments. We first present the dataset and data preprocessing; we then detail the experiments that were conducted on the training strategy and inclusion of infrared data.

\subsection{Dataset}
We use the TC-PRIMED v01r01 dataset \cite{razinTropicalCyclonePrecipitation2023}, which is publicly available. We extract microwave images from 11 microwave sensors (AMSR2 GCOMW1, TMI TRMM, GMI GPM, SSMI F11-13-14-15, and SSMIS F16-F19). We use the near-37GHz and near-89GHz channels, with both horizontal and vertical polarizations for each. This choice of frequencies has two motivations: their combination gives valuable information onto precipitation within the cyclone \cite{viltardEvaluationDrainDeeplearning2023,sambathUnsupervisedDomainAdaptation2024}, and they are available on many satellites across the GPM constellation. We also extract near-11µm infrared observations, which are pre-centered around the cyclones' centers in TC-PRIMED.  We use observations spanning from 1987 to 2024, of which the 2005-07-14-21-23 seasons are reserved for validation, 2006-08-15-22-24 for testing, and all other seasons are used for training. The dataset includes images over all major basins.


Each source is normalized individually by subtracting the mean of each channel and dividing its by its standard deviation. We add a land-sea mask obtained via the \code{global-land-mask} Python package. The characteristics (IFOV and frequency) described in Section \ref{sec:source_item} are each normalized jointly across all microwave sources by subtracting the minimum and dividing by the maximum minus the minimum.

\subsection{Training strategies}
\label{sec:training_strategies}
In this section, we describe two strategies to train the model. These differ in the way the training samples are defined, and by which source may be used as target. Both strategies rely on a hyperparameter $\Delta t_{max}$, as explained in \ref{sec:problem}, which is the maximum time difference between two input sources. In this sense, $\Delta t_{max}$ can be thought of as an assimilation window.

\subsubsection{Supervised strategy}
\begin{figure}
    \centering
    \includegraphics[width=1.0\linewidth]{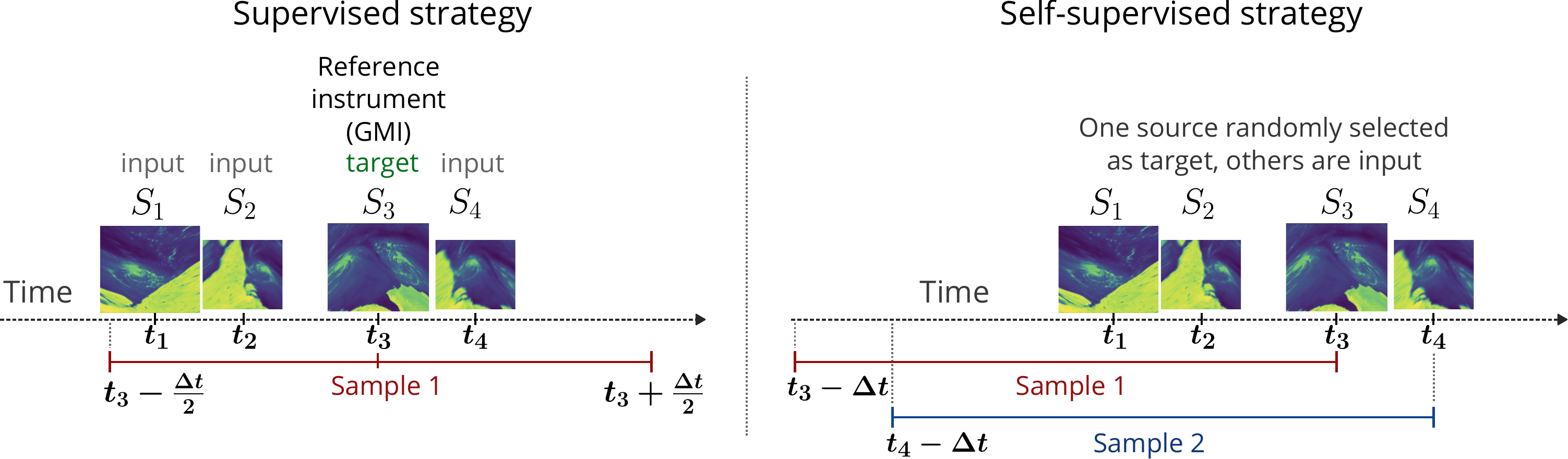}
    \caption{Scheme of the training strategies. Left: supervised strategy, in which the samples are defined as time windows centered around GMI observations. Right: self-supervised strategy, where every observation in the dataset (not only GMI) defines a sample. In this case, the target source is randomly selected among all sources within the time window.}
    \label{fig:training_strategies}
\end{figure}

Since our objective is to learn a model that generates images from the GMI instrument, a natural way to train the model is to train in a supervised manner, using GMI images as target and all other sources as inputs. In this context, we define the samples by the following process, schematized in Figure \ref{fig:training_strategies}: each GMI observation $\bm{x}_0$ in the dataset defines a time window $\omega = [t_0-\frac{\Delta t_{max}}{2}, t_0+\frac{\Delta t_{max}}{2}]$ centered around the time $t_0$ of the GMI image. Then, all observations $\bm{x}_i$ of the same cyclone in the dataset from sources other GMI that are within $\omega$ are gathered into a set of input sources $(\bm{x}_1,\bm{x}_2,...,\bm{x}_K)$. The model is then tasked with reconstructing the GMI image $\bm{x}_0$, conditioned on $(\bm{x}_1,\bm{x}_2,...,\bm{x}_K)$. We only consider samples for which at least one input source is available, i.e. there is at least one image from another source that is temporally separated from the GMI image by $\frac{\Delta t_{max}}{2}$ or less. This strategy results in 5,876 training samples.

\subsubsection{Self-supervised strategy}
While the supervised strategy is a natural approach, the quantity of training samples is limited to the number of GMI images which are close enough in time to an observation of the same cyclone from another source. For this reason, we use a self-supervised strategy proposed in \cite{dauvilliersMoTiFSelfsupervisedModel2025}, to increase the amount of training data. The process is the following: every observation $\bm{x}$ in the dataset (as opposed to GMI images only) defines a time window $\omega = [t_0-\Delta t_{max}, t_0]$. Then again, all observations of the same cyclones that are within $\omega$ are gathered to form a sample $\bm{X}=(\bm{x}_0,\bm{x}_1,\bm{x}_2,...,\bm{x}_K)$. Instead of always using the reference image $\bm{x}_0$ as target as in the supervised strategy, one of the observations $(\bm{x}_0,\bm{x}_1,\bm{x}_2,...,\bm{x}_K)$ is randomly selected as the target with uniform probability. 

This augments the training data in two ways compared to the supervised strategy: first, we keep all samples for which at least any two sources are temporally separated by at most $\Delta t_{max}$, instead of necessarily including a GMI image. This raises the number of training samples from 5,876 with the supervised strategy to 47,793. While most of these samples do not include a GMI observation, which is the source that we eventually aim to interpolate and that we evaluate our model on, we make the hypothesis that the tasks are still correlated enough that the model will benefit from the gain in training samples. Second, the random target selection means that for a given sample, the task differs between two training epochs, virtually further increasing the diversity of the training task.

\subsection{Impact of including infrared data in the input sources}
After assessing the impact of the training strategy, we also compare the performances of two models respectively trained with and without including infrared images in the input sources. As, to our knowledge, no previous work has used both microwave and infrared to interpolate microwave images of tropical cyclones, the aim is to assess the usefulness of combining these two modalities. This comparison is done on models trained with the self-supervised strategy. The training parameters used across the experiments are listed in Appendix 1.

\section{Results}
\label{sec:results}

\subsection{Evaluation metrics}
\paragraph{Continous Ranked Probability Score (CRPS)} The CRPS measures the quality of an ensemble of predictions against a single observation, and is only commonly used in numerical weather forecasting to evaluate probabilistic models \cite{priceProbabilisticWeatherForecasting2025,couaironArchesWeatherArchesWeatherGenDeterministic2024}. It rewards both sharpness (error in the ensemble mean) and calibration, by comparing the predicted distribution with the marginal single-step distribution defined by the single true observation:
\begin{equation*}
    \text{FairCRPS}((y^\theta_1,y^\theta_2,...y^\theta_M),y) = \frac{1}{M}\sum_{i=1}^M|y^\theta_i-y| - \frac{1}{2M(M-1)}\sum_{i,j=1}^M|y^\theta_i-y^\theta_j|
\end{equation*}
where $M$ is the ensemble size. We note that we use the Fair version of the CRPS, which is unbiased for finite ensembles \cite{zamoEstimationContinuousRanked2018}, as we only use $M=10$ members. The CRPS is computed at each pixel, then averaged over the image.

\paragraph{Skill-Spread ratio}
The Skill-Spread Ratio (SSR) measures the dispersion of an ensemble as the ratio between the average standard deviation of the members and the ensemble mean RMSE:
\begin{equation*}
    \text{SSR} = \sqrt{\frac{M}{M+1}} \frac{\sigma_{\text{spread}}}{\text{RMSE}_{\text{ensemble mean}}}
\end{equation*}
A well-calibrated ensemble should have an SSR close to 1 \cite{fortinWhyShouldEnsemble2014}. An SSR below 1 means the ensemble is underdispersive, while an SSR above 1 means it is overdispersive. We note that we use the $\frac{M+1}{M}$ correction as we use a relatively low number of members (10).

\paragraph{Deterministic metrics} We use the Root Mean Square Error (RMSE), Mean Absolute Error (MAE), and Mean Absolute Percentage Error (MAPE) and to compare the performances of a deterministic model versus the ensemble mean of our flow matching model.

\subsection{Evaluation setting}
\label{sec:evaluation_setting}
The evaluation samples are defined following the same process as in the supervised strategy: for all cyclones within the test seasons, we gather every available GMI image. For each such GMI image $\bm{x}_0$, we define a time window $\omega = [t_0-\frac{\Delta t_{max}}{2}, t_0+\frac{\Delta t_{max}}{2}]$, and gather all observations of the same storm that occur within $\omega$. These sources are used as input for the reconstruction of the GMI image, which is used as groundtruth to evaluate the metrics. For all flow matching models, 10 random realizations are generated for each sample, with the same starting noise points across models.

\subsection{Impacts of adding self-supervised training and infrared input}
\label{sec:results_fm}
In this section, we ablate over the following choices: (a) training with the supervised strategy (described in Section \ref{sec:training_strategies}), and including only microwave data as input sources; (b) training with the self-supervised strategy, still only with microwave inputs; and (c) training with the self-supervised strategy, including both microwave and infrared sources as input. For this evaluation, all models are trained with exactly the same architecture and training hyperparameters, and are evaluated over the same test samples.

\begin{figure}[!ht]
    \centering
    \begin{subfigure}[b]{0.45\textwidth}
        \centering
        \includegraphics[width=\textwidth]{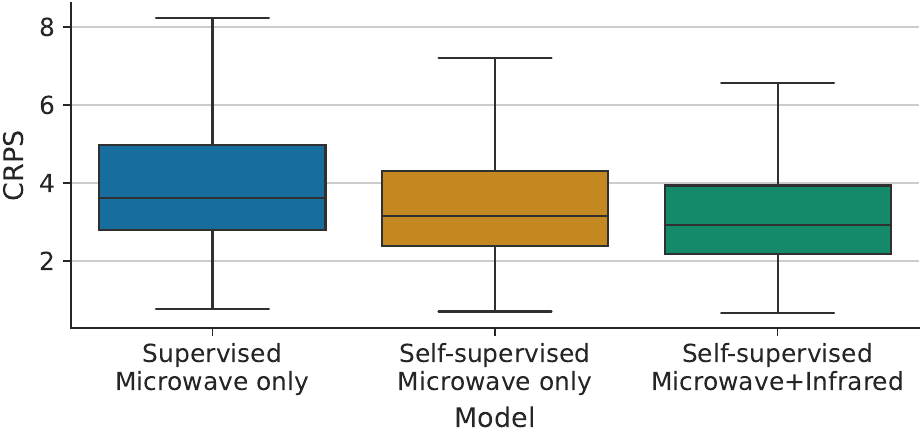}
        \caption{Distributions of CRPS on the test samples for three models with different training configurations and input sources. Lower is better.}
        \label{fig:crps}
    \end{subfigure}
    \hfill 
    \begin{subfigure}[b]{0.45\textwidth}
        \centering
        \includegraphics[width=\textwidth]{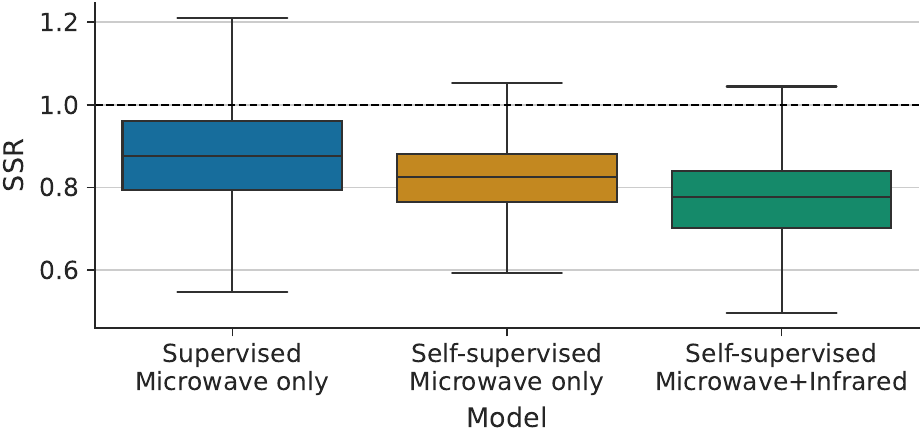}
        \caption{Distribution of Spread-Skill Ratio (SSR) on the test samples for each flow matching model. All models are below 1.0, indicating underdispersion.}
        \label{fig:ssr}
    \end{subfigure}

    \vspace{1em} 
    
    \begin{subfigure}[b]{0.45\textwidth} 
        \centering
        \includegraphics[width=\textwidth]{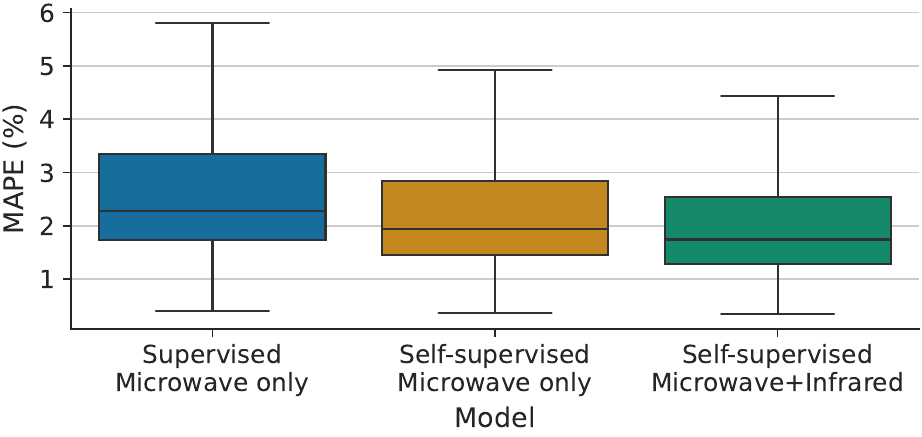}
        \caption{Distribution of the ensemble mean Mean Absolute Percentage Error (MAPE) on the test samples for each flow matching model.}
        \label{fig:mape}
    \end{subfigure}
    
    \caption{Comparison of CRPS, SSR and MAPE between flow matching models.}
    \label{fig:fm_results}
\end{figure}

The distributions of CRPS over the test samples for the three models are presented in Figure \ref{fig:crps}. Training the model with the self-supervised strategy over the supervised improves the average CRPS by 13.2\%, with a further improvement of 8\% gained by combining infrared and microwave data instead of microwave only. That conclusion is also valid according to the RMSE, MAE and MAPE, as displayed in Table \ref{tab:main_results}. On the other hand, Figure \ref{fig:ssr} displays the Spread-skill ratio for each model. While all models are underdispersive (SSR below 1.0), the addition of self-supervised training and infrared inputs worsen that caveat. We conclude from this that the gap in CRPS between the models stems from the improvement in accuracy of the ensemble mean (as shown in Figure \ref{fig:mape}), compensating the lower spread.

\subsection{Comparing generative models with a deterministic baseline}
In this section, we compare our best flow matching model (self-supervised, using both microwave and infrared) to a deterministic baseline. The deterministic model uses exactly the same architecture as the flow matching model, and is trained with the Root Mean Square Error (RMSE) as loss function. This deterministic baseline is an adaptation of the MoTiF architecture proposed in \cite{dauvilliersMoTiFSelfsupervisedModel2025}, which is to our knowledge the only model in the literature that can be applied to data with the level of heterogeneity faced in this work. Compared to the original MoTiF, the following changes have been applied: first, the multi-source cross-attention layer has been replaced with the cross-source attention layer described in \ref{sec:architecture}; and second, an embedding layer has been added to let the model use infrared observations as input.

\begin{table}[!ht]
  \centering
  \scriptsize
  \caption{Quantitative evaluation comparing the flow matching ensemble means and a deterministic baseline. The values are presented as the average metric over the test samples, followed by a 95\% confidence interval in brackets. The MAE and RMSE are in Kelvin. "MW" stands for Microwave, while "IR" stands for Infrared. Lower is better for all metrics.}
  \label{tab:main_results}
  \begin{tabularx}{1.0\linewidth}{X c c c c c c}
    \toprule
    \textbf{Model type} & \textbf{Training strategy} & \textbf{Input sources} & \textbf{CRPS} & \textbf{RMSE} & \textbf{MAE} & \textbf{MAPE (\%)} \\
    \midrule
    Deterministic & Self-supervised & MW+IR & - & \makecell{9.32 \\ {[9.19, 9.45]}} & \makecell{5.15 \\ {[5.07, 5.22]}} & \makecell{2.26 \\ {[2.22, 2.30]}} \\
    \addlinespace
    Flow matching & Supervised & MW only & \makecell{4.08 \\ {[4.02, 4.14]}} & \makecell{10.92 \\ {[10.77, 11.07]}} & \makecell{6.29 \\ {[6.19, 6.38]}} & \makecell{2.79 \\ {[2.73, 2.84]}} \\
    \addlinespace
    Flow matching & Self-supervised & MW only & \makecell{3.54 \\ {[3.49, 3.60]}} & \makecell{9.72 \\ {[9.58, 9.85]}} & \makecell{5.31 \\ {[5.23, 5.39]}} & \makecell{2.34 \\ {[2.30, 2.39]}} \\
    \addlinespace
    Flow matching & Self-supervised & MW+IR & \makecell{\textbf{3.26} \\ {[3.21, 3.31]}} & \makecell{\textbf{9.01} \\ {[8.89, 9.13]}} & \makecell{\textbf{4.76} \\ {[4.69, 4.83]}} & \makecell{\textbf{2.09} \\ {[2.05, 2.13]}} \\
    \bottomrule
  \end{tabularx}
\end{table}

\paragraph{Quantitative comparison} The quantitative comparison of the deterministic model against the flow matching models is presented in Table \ref{tab:main_results}. As the deterministic model only outputs a single prediction, we do not compare its CRPS, which inherently advantages probabilistic models. The deterministic metrics (RMSE, MAE and MAPE) are computed against the ensemble mean of the flow matching models. When using self-supervised training and including infrared in the inputs to both our flow matching model and the deterministic MOTIF baseline, the generative model produces an ensemble mean that is on par (and actually marginally better) than the deterministic baseline. This result validates that the switch to generative modeling does not come at the cost of a worse estimation of the conditional expectation.

\begin{figure}[!ht]
    \centering
    \includegraphics[width=1.0\linewidth]{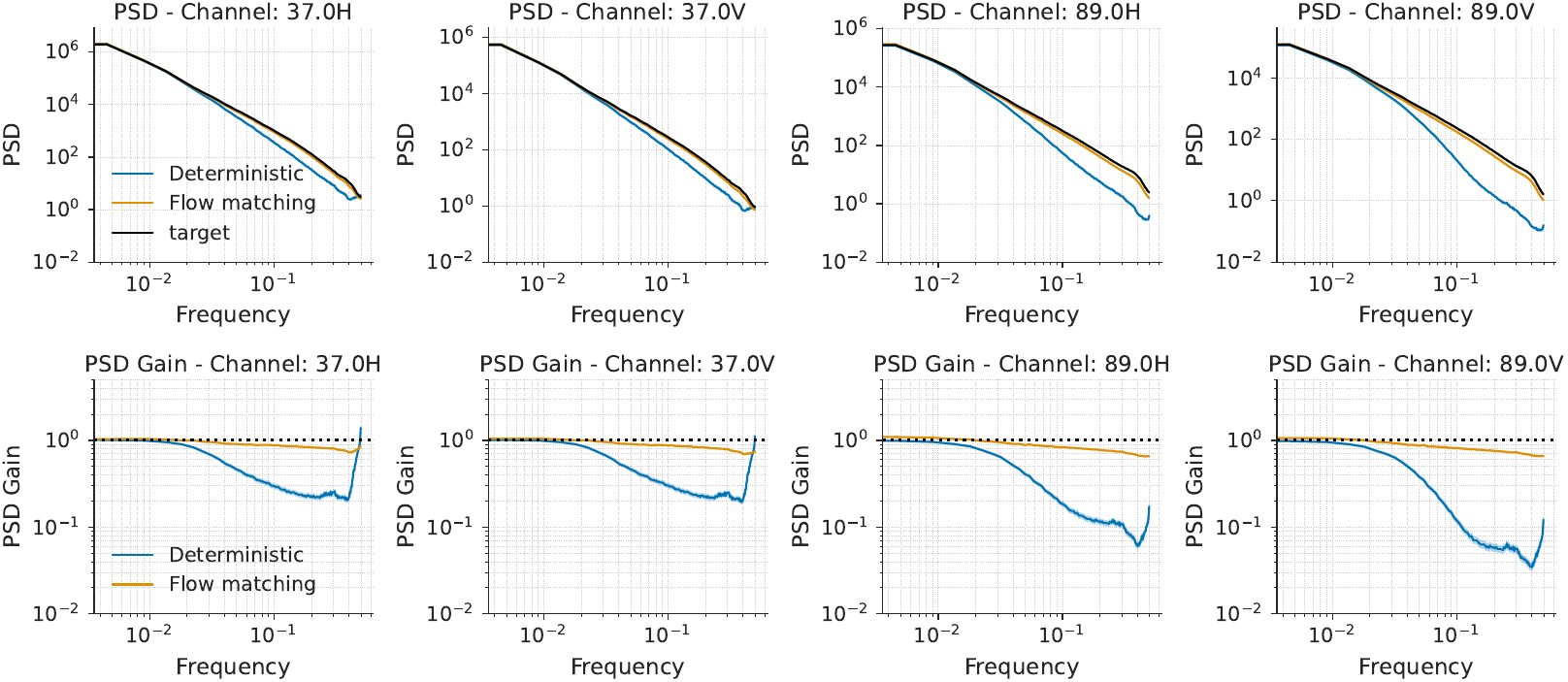}
    \caption{Comparison of the radially-averaged Power Spectral Density (PSD), averaged over all test samples, between a flow matching and a deterministic model (self-supervised, using both microwave and infrared as input). The PSD Gain is computed by dividing the predictions' PSD by that of the target at each frequency. We attribute the spikes at the very highest frequencies in the deterministic spectrum to artifacts due to the patching process.}
    \label{fig:spectrum}
\end{figure}

\paragraph{Power spectrum comparison} The main aim of training a generative model is its supposed ability to generate realistic predictions, notably by producing a power spectrum close to that of true data. Mid and high frequencies are especially important in the context of tropical cyclones microwave images, as extreme precipitation and winds occur in small-scale spatial features. In this regard, the flow matching model reproduces the target spectrum more accurately, as shown in Figure \ref{fig:spectrum}. The mid and high frequencies are particularly better conserved for the 89GHz channels, which have a higher spatial frequency than the 37GHz ones. Additional visualizations are included in Appendix \ref{sec:additional_visualizations}.

\section{Discussion and Conclusion}
\label{sec:conclusion}
To summarize, this work proposes the first generative model to interpolate microwave images of tropical cyclones conditioned on data from multiple microwave and infrared instruments. To deal with the high heterogeneity of the data, we use an architecture specifically designed for spatio-temporally misaligned geospatial sources. We show that our model benefits from being trained in a self-supervised manner, as well as combining both infrared and microwave data. Using these benefits, our model retains an ensemble mean accuracy comparable to that of a deterministic baseline, while better reproducing high-frequency features.

One of the direct directions of improvement is to correct the underdispersivity, which is increased by self-supervised training and using infrared data. A plausible way to increase the spread is to calibrate the input noise variance, by multiplying the noise vector by a constant factor. Besides, the self-supervised model presented in this work is not fine-tuned on GMI targets specifically, meaning it is calibrated to fit the spread of the full ensemble of microwave sensors used during training.

At longer term, future work could include for example looking at other types of weather events such as extreme precipitation. Besides, while this is to our knowledge the only deep learning architecture that can be applied to such heterogeneous data, there could most definitely be other approaches to test in the future. Still, we hope this work can serve as inspiration for further works on highly heterogeneous geospatial data.

\vspace{1em}
\textbf{Reproducibility} The data used is publicly available through \href{https://rammb-data.cira.colostate.edu/tcprimed/}{the TC-PRIMED website}.The code is available at \url{https://github.com/dauvillc/motif}.

\textbf{Use of Generative AI} Generative AI was used to format the tables and figures, as well as for coding. AI was not used to write or rephrase content in this paper.

\section{Author contributions}
C.D.: Conceptualization (lead), Methodology (lead), Software, Validation, Investigation, Data Curation, Writing - Original Draft, Visualization.\\
C.M.: Conceptualization (support), Methodology (support), Writing - Review \& Editing, Supervision, Project Administration, Resources, Funding Acquisition.

\section{Acknowledgments}
We thank Nicolas Viltard, Cécile Mallet and Laurent Barthès for sharing their knowledge regarding microwave imagery and tropical cyclones. Many thanks to Anastase Charantonis for the fruitful discussions regarding the methodology. C.D. and C.M. were supported by the French government, via the Choose France Chair in AI. This work was granted access to the HPC resources of IDRIS under the allocation AD011014682R2 made by GENCI.

%
%
%
\bibliographystyle{splncs04}
\bibliography{motif_gen}

\newpage
\appendix
\vspace{2em} 
\begin{center}
    \Huge \textbf{Appendix} 
\end{center}
\vspace{1em} 
\addcontentsline{toc}{section}{Appendix}

\section{Training details}
\label{ref:training_details}
Table 1 list the architectural and optimization parameters used for all flow matching and deterministic experiments.
\begin{table}
  \centering
  \scriptsize
  \caption{Training settings used for the experiment.}
  \label{tab:training_settings}
  \begin{tabularx}{\linewidth}{>{\raggedright\arraybackslash}X l >{\raggedright\arraybackslash}X l}
    \toprule
    \textbf{Architecture} &  & \textbf{Optimization} &  \\
    \midrule
    Number of blocks in the backbone & 12 & Optimizer & AdamW \\
    \addlinespace
    Pixels embedding dim. $d_{\text{pixels}}$ & 512 & $\beta_1, \beta_2$ & 0.9, 0.999 \\
    \addlinespace
    Coordinates embedding dim. $d_{\text{coords}}$ & 256 & Weight decay & 0.05 \\
    \addlinespace
    Conditioning embedding dim. $d_{\text{cond}}$ & 256 & LR Schedule & Cosine annealing \\
    \addlinespace
    Number of attention heads & 8 & Min LR, Max LR & $10^{-6}, 3\times 10^{-4}$ \\
    \addlinespace
    Cross-source att. window size & 4 & LR warmup & \makecell[l]{1 epoch, linear \\ from 0 to $3\times 10^{-4}$} \\
    \addlinespace
    Source-wise att. window size & 8 & Epochs & 100 \\
    \addlinespace
    MLP inner expansion ratio & 2.0 & Checkpoint selection & Best val. loss \\
    \bottomrule
  \end{tabularx}
\end{table}

\section{Impact of sources availability}
\label{sec:multiple_sources}

\begin{figure}[!ht]
    \centering
    \begin{subfigure}[b]{0.7\textwidth}
        \centering
        \includegraphics[width=\textwidth]{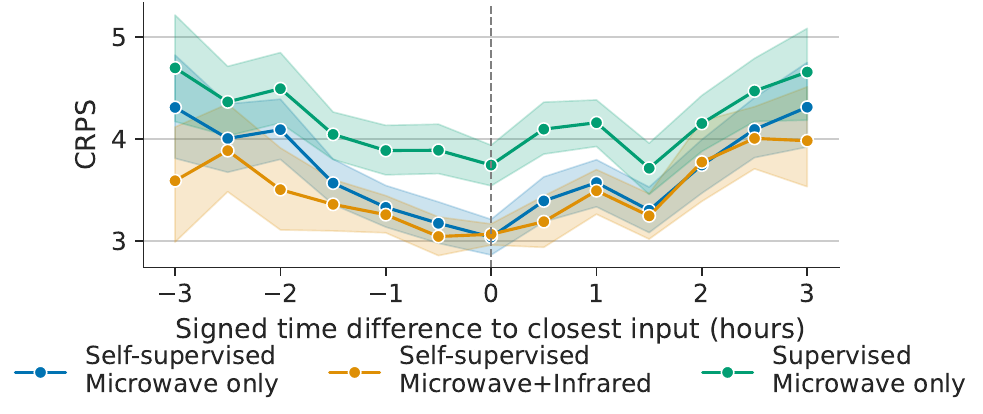}
        \label{fig:crps_vs_min_dt}
    \end{subfigure}
    \vspace{1em}
    \begin{subfigure}[b]{0.7\textwidth}
        \centering
        \includegraphics[width=\textwidth]{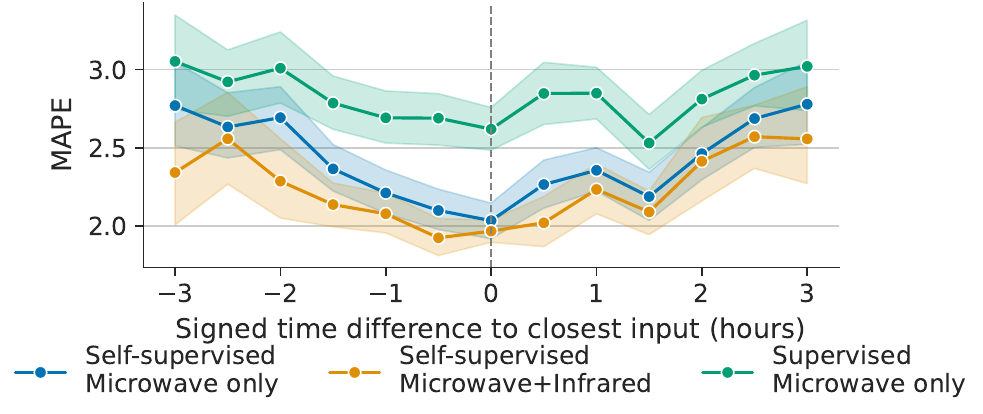}
        \label{fig:mape_vs_min_dt}
    \end{subfigure}
    \vspace{1em}
    \begin{subfigure}[b]{0.7\textwidth}
        \centering
        \includegraphics[width=\textwidth]{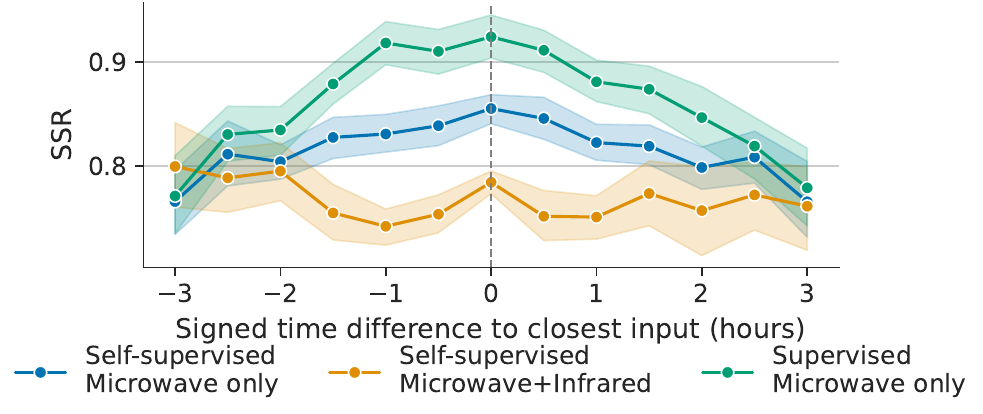}
        \label{fig:ssr_vs_min_dt}
    \end{subfigure}\vspace{1em}
    \begin{subfigure}[b]{0.7\textwidth}
        \centering
        \includegraphics[width=\textwidth]{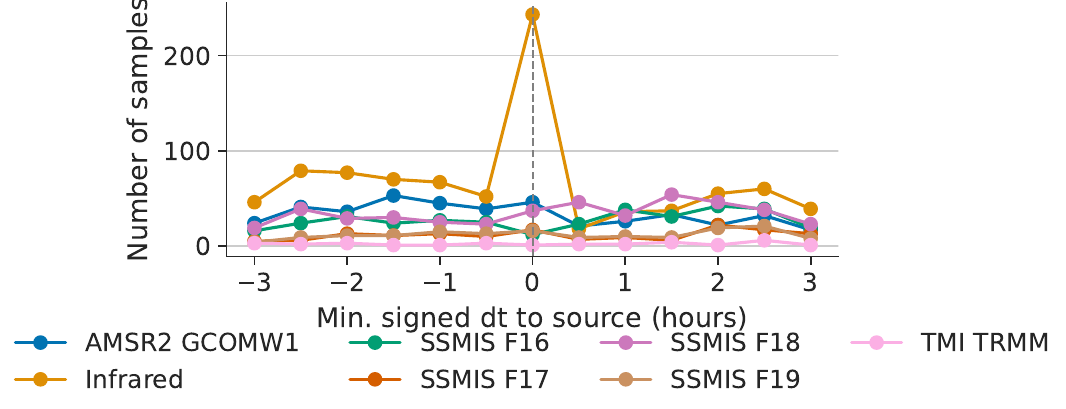}
        \label{fig:counts_vs_min_dt}
    \end{subfigure}
    \caption{CRPS, MAPE and Skill-Spread ratio against time delta (dt) between the target and the source that is temporally closest to it. THe time delta is binned in 30min intervals, centered on dt=0. Shaded areas indicate 95\% confidence intervals.}
    \label{fig:vs_min_dt}
\end{figure}

\begin{figure}[!ht]
    \centering
    \begin{subfigure}[b]{0.7\textwidth}
        \centering
        \includegraphics[width=\textwidth]{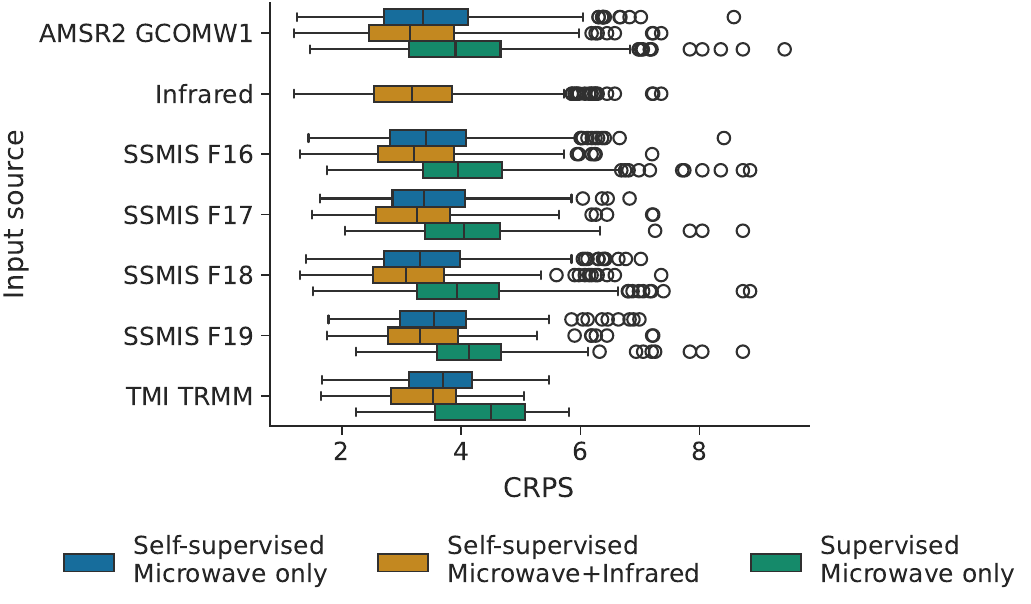}
        \label{fig:crps_per_source}
    \end{subfigure}
    \vspace{1em}
    \begin{subfigure}[b]{0.7\textwidth}
        \centering
        \includegraphics[width=\textwidth]{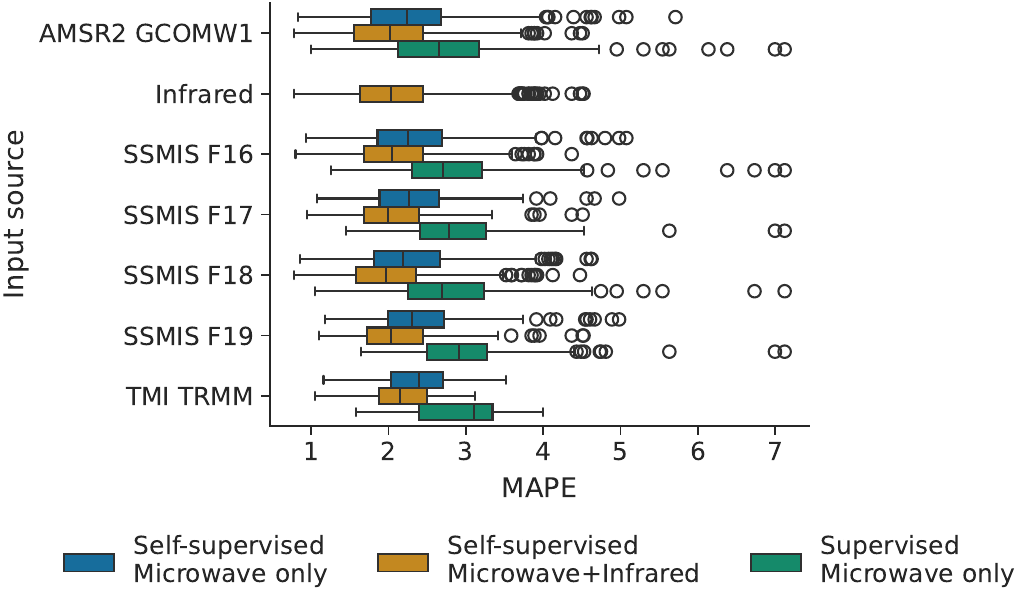}
        \label{fig:mape_per_source}
    \end{subfigure}
    \vspace{1em}
    \begin{subfigure}[b]{0.7\textwidth}
        \centering
        \includegraphics[width=\textwidth]{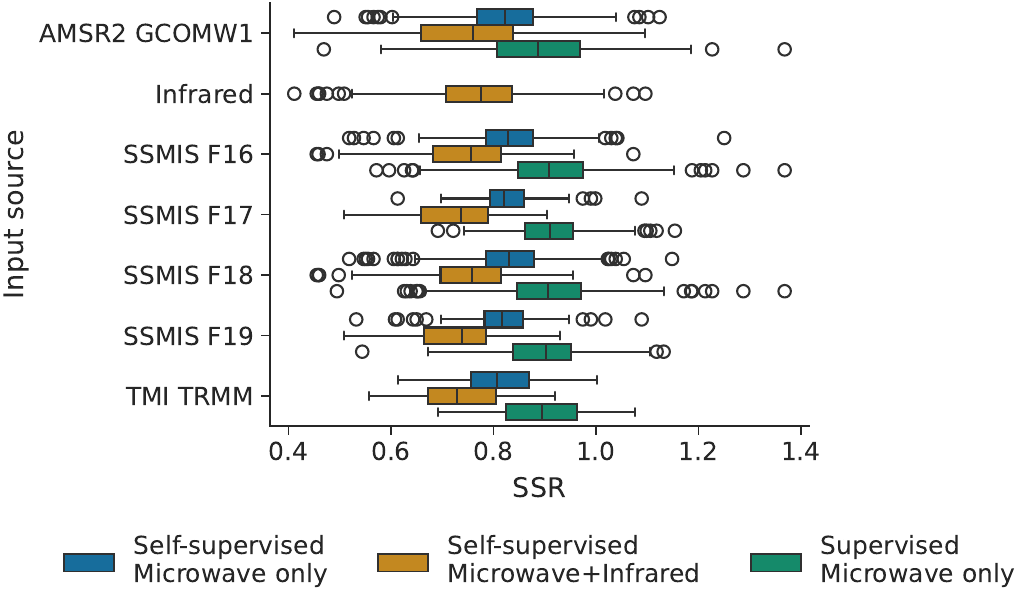}
        \label{fig:ssr_per_source}
    \end{subfigure}
    \vspace{1em}
    \begin{subfigure}[b]{0.7\textwidth}
        \centering
        \includegraphics[width=\textwidth]{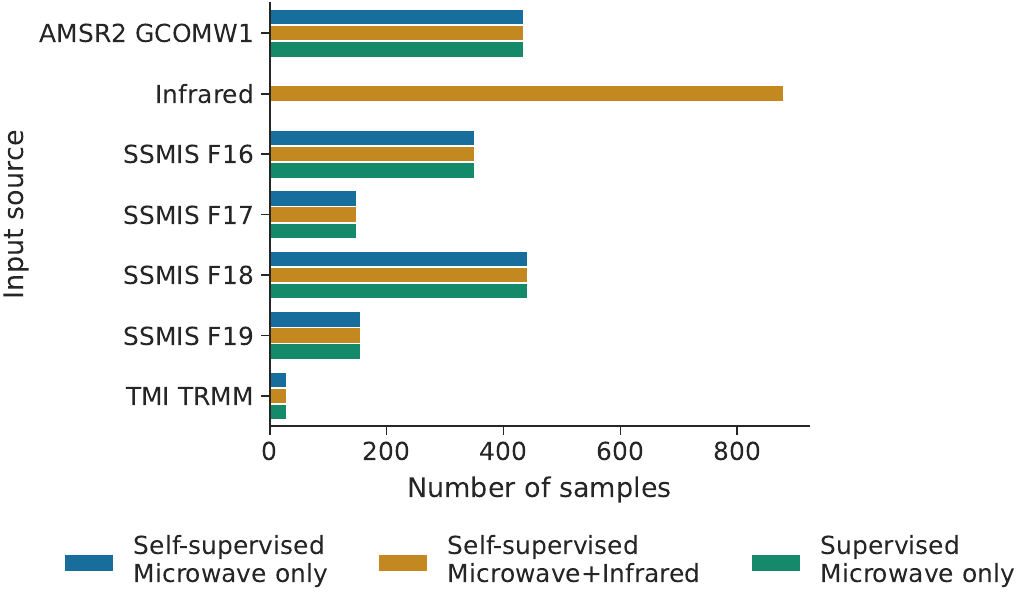}
        \label{fig:counts_per_source}
    \end{subfigure}
    \caption{For each input source, values of CRPS, MAPE and Skill-Spread ratio averaged over the test samples containing that source.}
    \label{fig:per_source}
\end{figure}

This section displays the variation in performance depending on which input sources are available. Figure \ref{fig:vs_min_dt} shows the variations in CRPS, MAPE and SSR depending on the time between the target image and the input source that's closest to it in time. We observe that, as expected, the metrics generally improve as the model gets a source that's closer in time to its target. Figure \ref{fig:per_source} displays the metrics marginalized over each input source specifically.

\newpage

\section{Additional visualizations}
\label{sec:additional_visualizations}

\begin{figure}[!ht]
    \centering
    \includegraphics[width=1.0\linewidth]{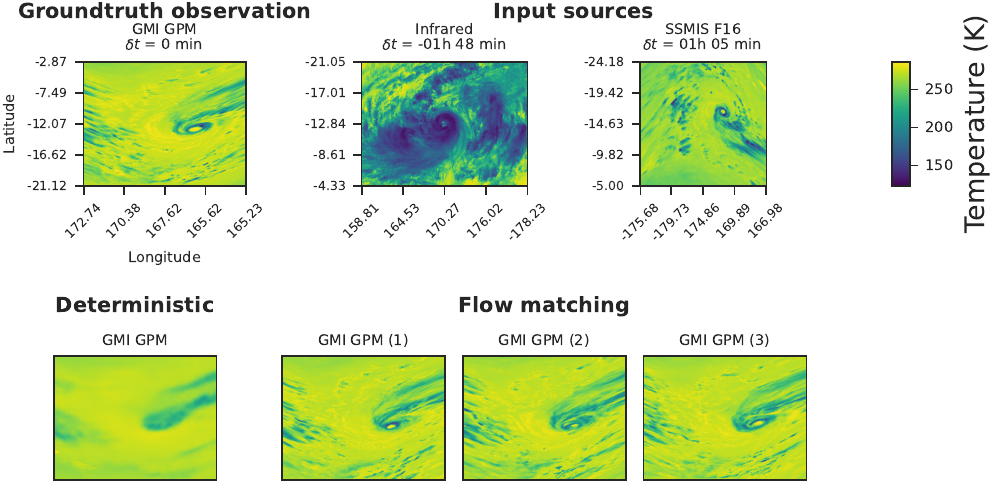}
\end{figure}

\begin{figure}[!ht]
    \centering
    \includegraphics[width=1.0\linewidth]{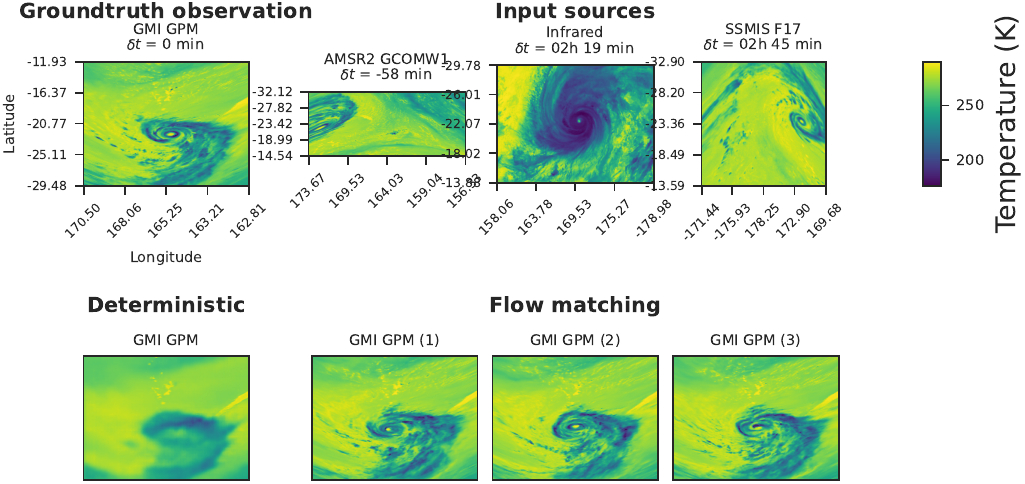}
\end{figure}

\begin{figure}[!ht]
    \centering
    \includegraphics[width=1.0\linewidth]{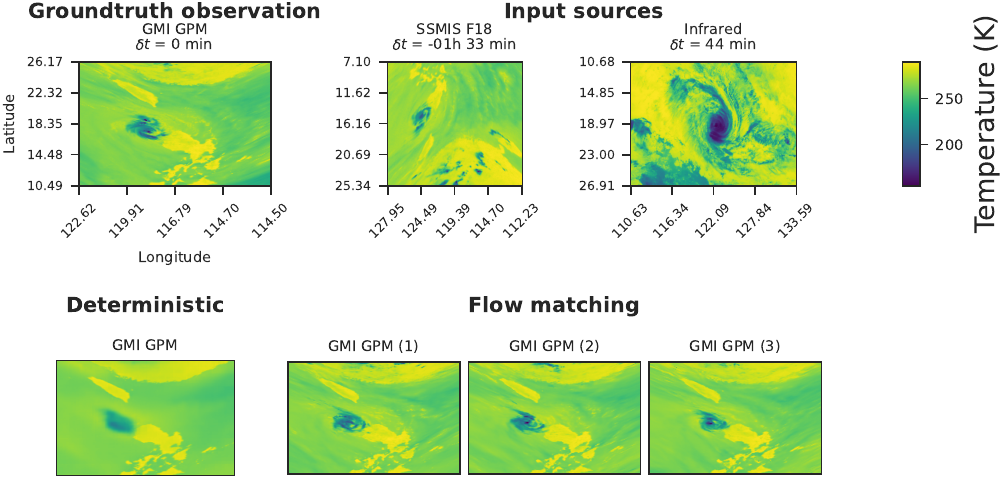}
\end{figure}

\begin{figure}[!ht]
    \centering
    \includegraphics[width=1.0\linewidth]{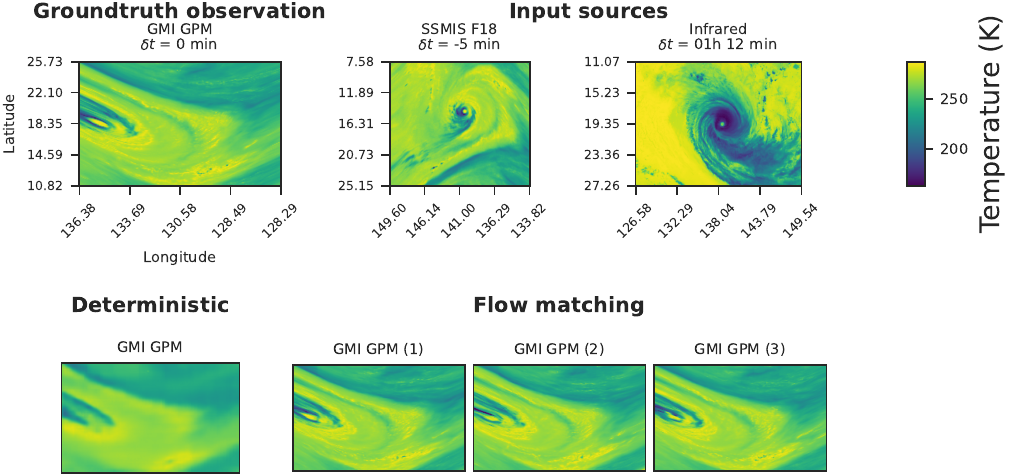}
\end{figure}

\begin{figure}[!ht]
    \centering
    \includegraphics[width=1.0\linewidth]{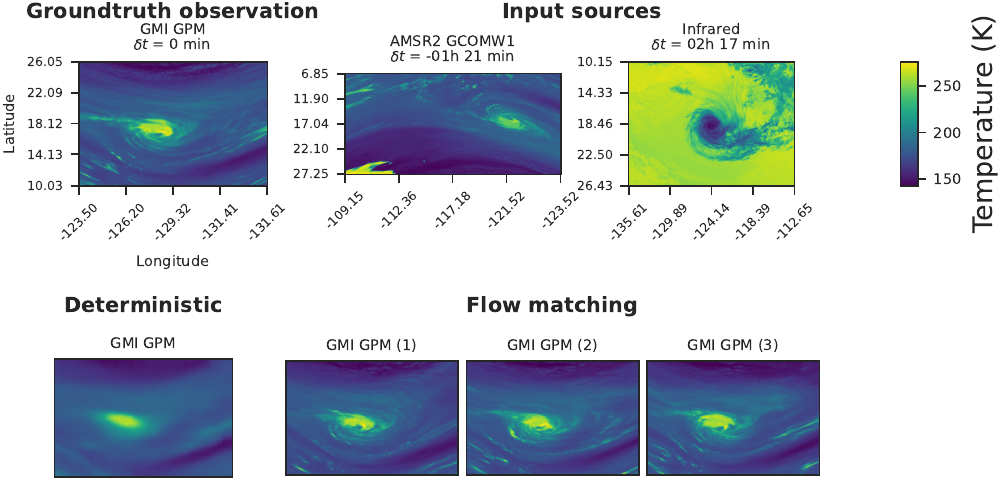}
\end{figure}

\begin{figure}[!ht]
    \centering
    \includegraphics[width=1.0\linewidth]{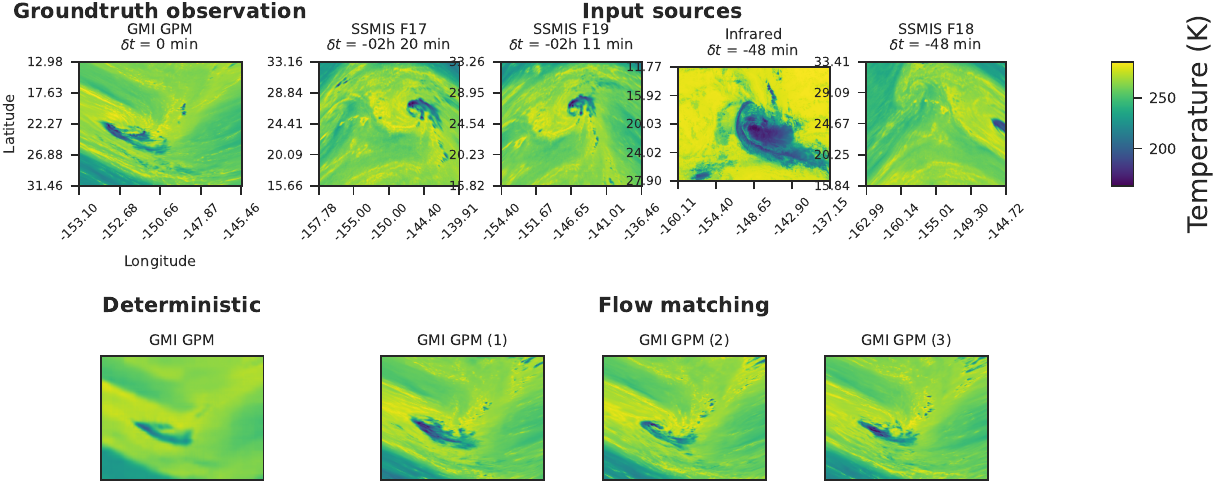}
\end{figure}

\begin{figure}[!ht]
    \centering
    \includegraphics[width=1.0\linewidth]{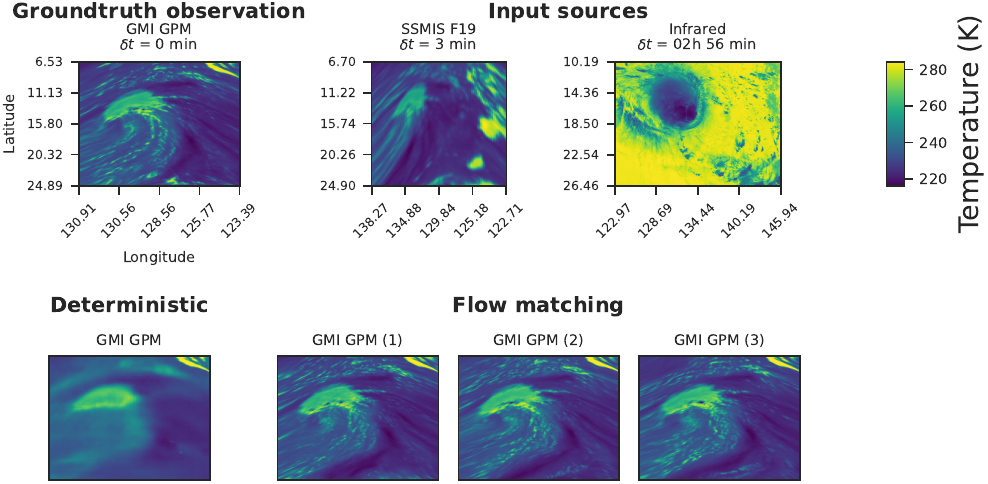}
\end{figure}

\begin{figure}[!ht]
    \centering
    \includegraphics[width=1.0\linewidth]{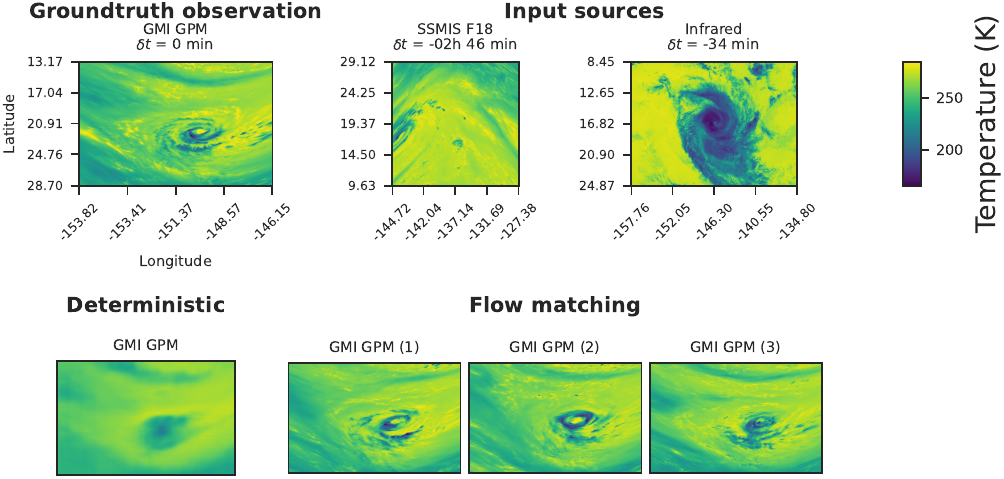}
\end{figure}

\begin{figure}[!ht]
    \centering
    \includegraphics[width=1.0\linewidth]{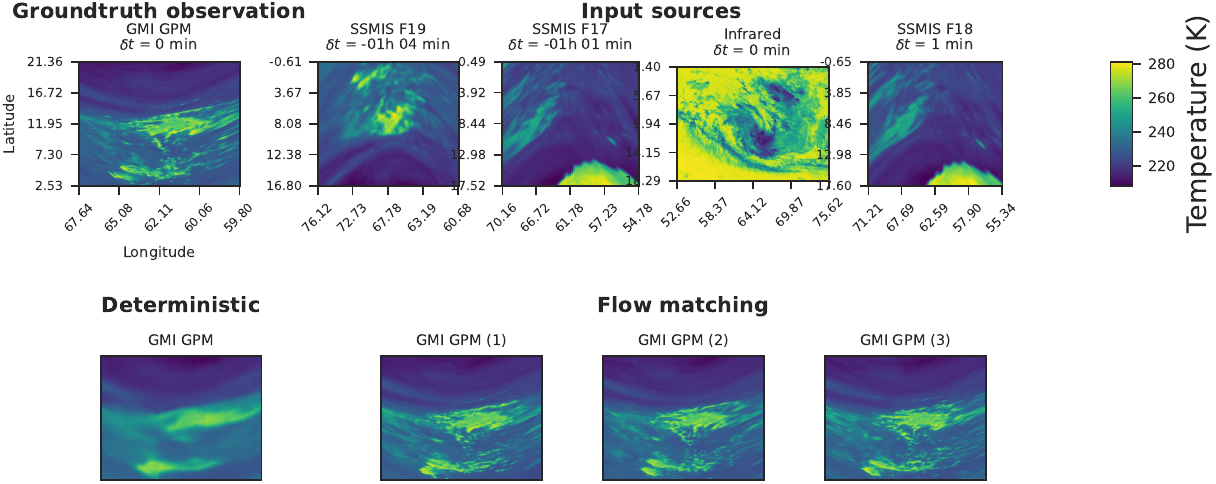}
\end{figure}

\end{document}